\documentclass{article}

\usepackage{PRIMEarxiv}

\usepackage[utf8]{inputenc} % allow utf-8 input
\usepackage[T1]{fontenc}

\usepackage[unicode=true,pagebackref=true,
colorlinks,linkcolor=blue,citecolor=green,final]{hyperref}       % hyperlinks
\usepackage{url}            % simple URL typesetting
\usepackage{booktabs}       % professional-quality tables
\usepackage{amsfonts}       % blackboard math symbols
\usepackage{nicefrac}       % compact symbols for 1/2, etc.
\usepackage{microtype}      % microtypography
\usepackage{lipsum}
\usepackage{fancyhdr}       % header
\usepackage{graphicx}       % graphics
\graphicspath{{media/}}     % organize your images and other figures under media/ folder
\usepackage{subcaption}
\usepackage{multirow}
\usepackage[super]{nth}
\title{Rethinking Gradient Weight’s Influence over Saliency Map Estimation}

\author{
 Masud An Nur Islam Fahim \\
  Chosun University\\
  Gwangju, South Korea \\
  \texttt{mostofafahim21@gmail.com} \\
   \And
 Nazmus Saqib \\
  Chosun University\\
  Gwangju, South Korea \\
  \texttt{nsaqib1995@gmail.com} \\
  \And
 Shafkat Khan Siam \\
 Chosun University\\
  Gwangju, South Korea \\
  \texttt{shafkat.kh022@gmail.com} \\
  \AND
  Ho Yub Jung\\
  Chosun University\\
  Gwangju, South Korea \\
  \texttt{jung.ho.yub@gmail.com}\\
}

\begin{document}
\maketitle

\begin{abstract}
Class activation map (CAM) helps to formulate saliency maps that aid in interpreting the deep neural network's prediction. Gradient-based methods are generally faster than other branches of vision interpretability and independent of human guidance. The performance of CAM-like studies depends on the governing model's layer response, and the influences of the gradients. Typical gradient-oriented CAM studies rely on weighted aggregation for saliency map estimation by projecting the gradient maps into single weight values, which may lead to over generalized saliency map. To address this issue, we use a global guidance map to rectify the weighted aggregation operation during saliency estimation, where resultant interpretations are comparatively clean er and instance-specific. We obtain the global guidance map by performing elementwise multiplication between the feature maps and their corresponding gradient maps. To validate our study, we compare the proposed study with eight different saliency visualizers. In addition, we use seven commonly used evaluation metrics for quantitative comparison. The proposed scheme achieves significant improvement over the test images from the ImageNet, MS-COCO 14, and PASCAL VOC 2012 datasets.

\end{abstract}

% keywords can be removed

\section{Introduction}

Deep neural networks have achieved superior performance on numerous vision tasks \cite{Zisserman1,Zisserman2,semanticsegmentation}. However, they contain complicated black boxes with huge, unexplainable parameters that begin with random initialization and reach another unpredictable (still) sub-optimal point. Such transformations are non-linear for each problem, the interpretability remains unsolved.

The vision community relies on estimating saliency maps for deciphering the decision-making process of deep networks. We try to bridge the unknown gap between the input space and decision space through this saliency map. These saliency map generation approaches can be divided into different procedures regarding input space \cite{RISE,SISE_main,Ada-SISE}, feature maps \cite{CAM,Axiom-CAM,Grad-CAM,Grad-CAM++,FullGrad,integrated,Ablation-CAM}, or propagation scheme \cite{Guidedbackpropagation,DEEPLIFT,LRP,pointinggame,Thereandbackagain}. Usual perturbation-based methods 
\cite{zeiler,externalperturbation1,externalperturbation2} probe the input space into different deviated versions and obtain a unified saliency map by their underlying algorithms. Even though their approach produces a better result, their process is relatively slow. CAM-like methods rely on the gradient for the given class and create a decent output within a very brief computation time.

\begin{figure*}[!htbp]
\centering
\begin{subfigure}[l]{0.80\textwidth}
    \centering
    \begin{subfigure}[t]{0.15\textwidth}
        \centering
        \includegraphics[height=15mm,width=\textwidth]{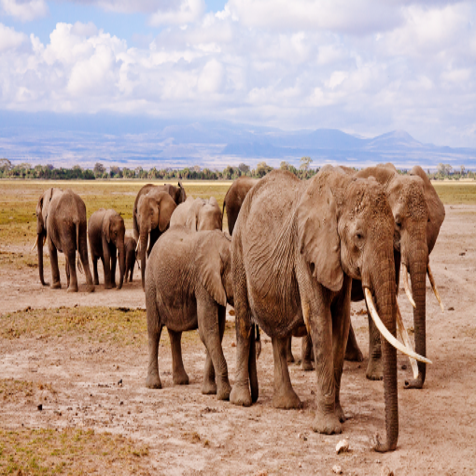}
       \subcaption{Single Class}
        \label{a}
    \end{subfigure}
    \begin{subfigure}[t]{0.15\textwidth}
        \centering
        \includegraphics[height=15mm,width=\textwidth]{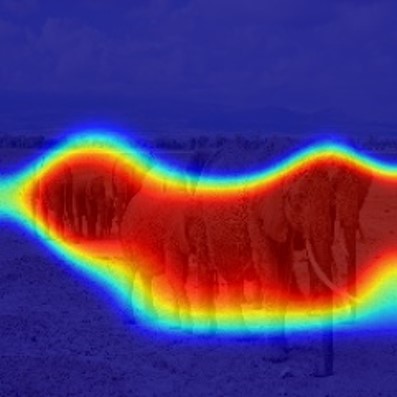}
       \subcaption{Feature map}
        \label{b}
    \end{subfigure}
    \begin{subfigure}[t]{0.15\textwidth}
        \centering
        \includegraphics[height=15mm,width=\textwidth]{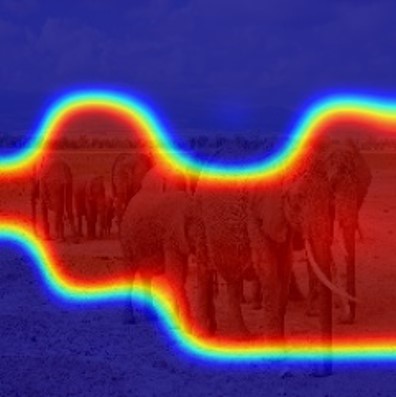}
       \subcaption{Ours}
        \label{c}
    \end{subfigure}
    \begin{subfigure}[t]{0.15\textwidth}
        \centering
        \includegraphics[height=15mm,width=\textwidth]{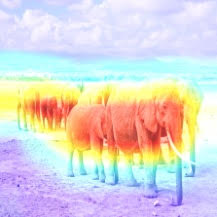}
       \subcaption{\cite{Score-CAM}}
        \label{d}
    \end{subfigure}
    \begin{subfigure}[t]{0.15\textwidth}
        \centering
        \includegraphics[height=15mm,width=\textwidth]{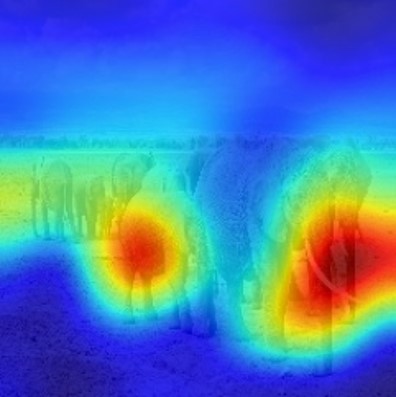}
       \subcaption{\cite{Grad-CAM}}
        \label{e}
    \end{subfigure}
    \begin{subfigure}[t]{0.15\textwidth}
        \centering
        \includegraphics[height=15mm,width=\textwidth]{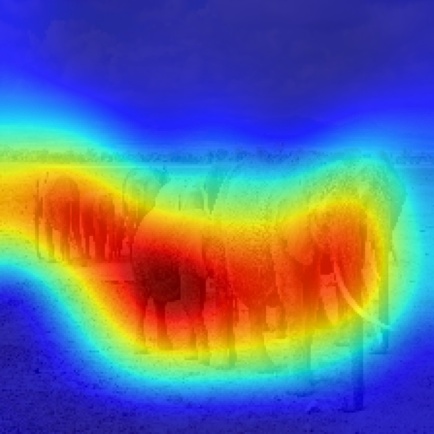}
       \subcaption{\cite{Grad-CAM++}}
        \label{f}
    \end{subfigure}
    \subcaption{Feature map characteristics for a single class}
    \label{sinef}
    \centering
\end{subfigure}

\begin{subfigure}[r]{0.90\textwidth}
    \centering
          \begin{subfigure}[t]{0.13\textwidth}
        \centering
        \includegraphics[height=13mm,width=\textwidth]{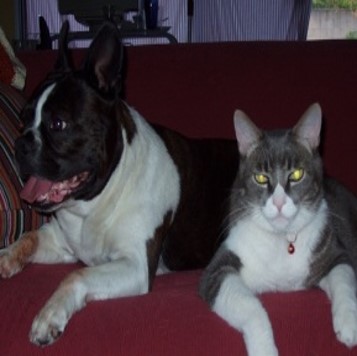}
       \subcaption{Dual Class}
        \label{g}
    \end{subfigure} 
          \begin{subfigure}[t]{0.13\textwidth}
        \centering
        \includegraphics[height=13mm,width=\textwidth]{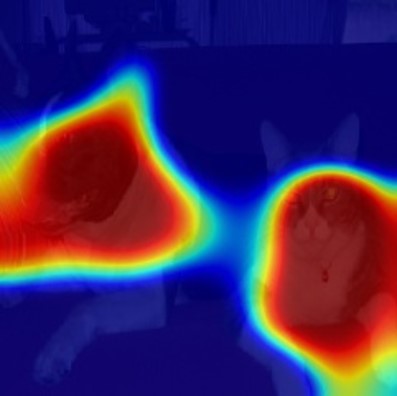}
       \subcaption{Feature map}
        \label{h}
    \end{subfigure}  
          \begin{subfigure}[t]{0.13\textwidth}
        \centering
        \includegraphics[height=13mm,width=\textwidth]{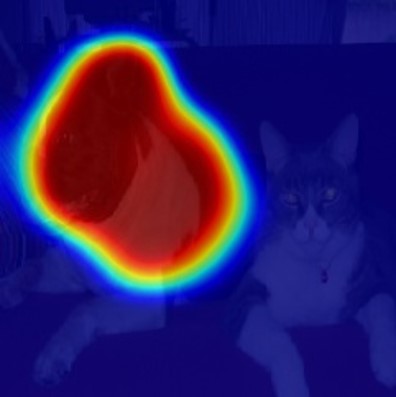}
       \subcaption{Ours(\nth{1})}
        \label{i}
    \end{subfigure}  
          \begin{subfigure}[t]{0.13\textwidth}
        \centering
        \includegraphics[height=13mm,width=\textwidth]{fig1/fifth_im.jpg}
       \subcaption{Ours(\nth{2})}
        \label{j}
    \end{subfigure}  
          \begin{subfigure}[t]{0.13\textwidth}
        \centering
        \includegraphics[height=13mm,width=\textwidth]{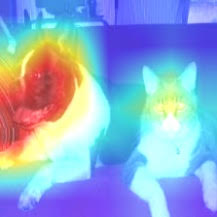}
       \subcaption{\cite{Score-CAM}}
        \label{k}
    \end{subfigure}
    \begin{subfigure}[t]{0.13\textwidth}
        \centering
        \includegraphics[height=13mm,width=\textwidth]{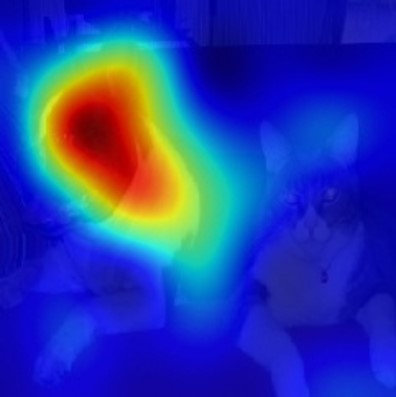}
       \subcaption{\cite{Grad-CAM}(\nth{1})}
        \label{l}
    \end{subfigure}
    \begin{subfigure}[t]{0.13\textwidth}
        \centering
        \includegraphics[height=13mm,width=\textwidth]{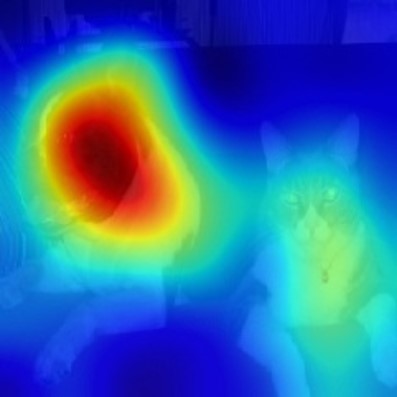}
       \subcaption{\cite{Grad-CAM++}(\nth{1})}
        \label{m}
    \end{subfigure} 
    \subcaption{Feature map characteristics for dual class}
    \label{duof}
\end{subfigure}
\caption{Proposed global guidance maps $g_M$ for cat, proposed saliency map ($S_c$) with and without global guidance, and the same maps for the dog. The proposed global guidance map provides strong localization information as well as exclusion of non-target class.}
\label{algorithm figure}
\end{figure*}

This study explores the established framework of CAM-based methods to address current issues within vision-based interpretability. Similar studies usually rely on gradient information to produce saliency maps. Some replace the gradient dependency with score estimation to build the saliency map out of the feature maps. Nonetheless, these studies formulate saliency maps that often suffer degradation during class discriminative examples. Moreover, their weighted accumulation does not always cover the expected region for the given single class instance, as the associated weights sometimes do not address the local correspondence. On the other hand, the gradient-less methods take significant time to produce a score index for the feature maps. 

Typically, the gradient maps are averaged into single values to produce the saliency map within CAM-like studies. However, this projection is not always efficient for the saliency map. In figure \ref{algorithm figure}, we can see that weighted multiplication still contains traces of unwanted classes. To address the weighted multiplication issue, we first look at the problem setup. For a given image, we use a model that produces a fixed number of feature maps before going into the dense layers. This fixation limits our search space for the saliency map. The only variables are the gradient maps that change according to the class. 

Gradient matrix to a single value conversion eventually reduces many significant gradients from influencing the overall accumulation. If we keep all the gradients and perform elementwise multiplication with feature maps, the obtained map contains better representations of the assigned class. The acquired map is the intermediate global guidance map for the usual local weighted multiplication. We use this guidance map to constrain the generated feature maps to be responsive only to the assigned class through element-wise matrix multiplication. After that, we perform the formal weighted accumulation for acquiring the saliency map, followed by carefully designed upscaling process. Thus, the produced saliency map is more responsive to the given class and its boundaries than previous approaches. The following lines summarize our overall contributions:

\begin{itemize}
\item A new saliency map generation scheme is formulated by introducing global guidance map that incorporates elementwise influence of the gradient tensor onto the saliency map.
\item Acquired boundaries for the saliency maps are crisper than contemporary studies and perform efficiently within single/multi-class/multi-instance-single class cases.
\item To validate our study, we perform seven different metric analyses on three different datasets, and the proposed study achieves state-of-the-art performance in most cases.
\end{itemize}

\section{Related Work}
\textbf{Backprop-based methods.}
Zisserman \textit{et al}.\cite{Zisserman1,Zisserman2} first introduced gradient calculation by focusing on computing the confidence score for  explanation generation, and other back propagation-based explanation studies have been introduced \cite{Guidedbackpropagation,integrated,FullGrad}. However, their gradient employment and manipulation lead to several issues, as addressed by \cite{integrated,FullGrad}. Instead of a single layer gradient, Srinivas \textit{et al.} \cite{FullGrad} focuses on aggregating gradients from all convolutional layers.

\textbf{Activation-based methods.}
Class activation mapping (CAM) methods are based on the study \cite{CAM}, where authors select feature maps as a medium for creating an explanation map. GradCAM \cite{Grad-CAM} is a weighted linear combination of the feature maps followed by the ReLU operation for a given image and the respective model. Later, GradCAM++ \cite{Grad-CAM++} introduced the effect of the higher-order gradients with inclusive only to positive elements properties. In this way, they achieve more precise representation compared to the previous studies. However, gradients are not the only way to generate the saliency map, which inspires the ScoreCAM \cite{Score-CAM}, AblationCAM \cite{Ablation-CAM}. X-GradCAM \cite{Axiom-CAM}, an extension to the GradCAM, follows the same the underlying weighted multiplication for GradCAM. EigenCAM \cite{Eigen-CAM} introduces principal component analysis into developing the saliency map. CAMERAS \cite{Cameras} utilizes basic CAM studies with multiscale input extension, leveraging the fusion technique of multiscale feature and gradient map weighted multiplication.

\textbf{Perturbation-based methods.}
Another group of studies treat the neural network as a ``white box" instead of a ``black box" and propose an explanation map by probing the input space. These studies generate saliency maps by checking the response upon the manipulated input space. By blocking/blurring/masking some of the regions randomly, these studies observe the forward pass response for each case and finally aggregate their decision to develop the saliency map. RISE \cite{RISE} first introduces such analysis, followed by external perturbation \cite{externalperturbation1,externalperturbation2}  where authors rely upon optimization procedure. SISE \cite{SISE_main} presents feature map selection from multiple layers, followed by attribution map generation and mask scoring to generate the saliency map. Later, they improvise it through adaptive mask selection in the ADA-SISE \cite{Ada-SISE} study.

\section{Proposed methodology}
\begin{figure*}[!htbp]\label{f4}
	\centering
	\includegraphics[width = 6 in, height= 2.5 in]{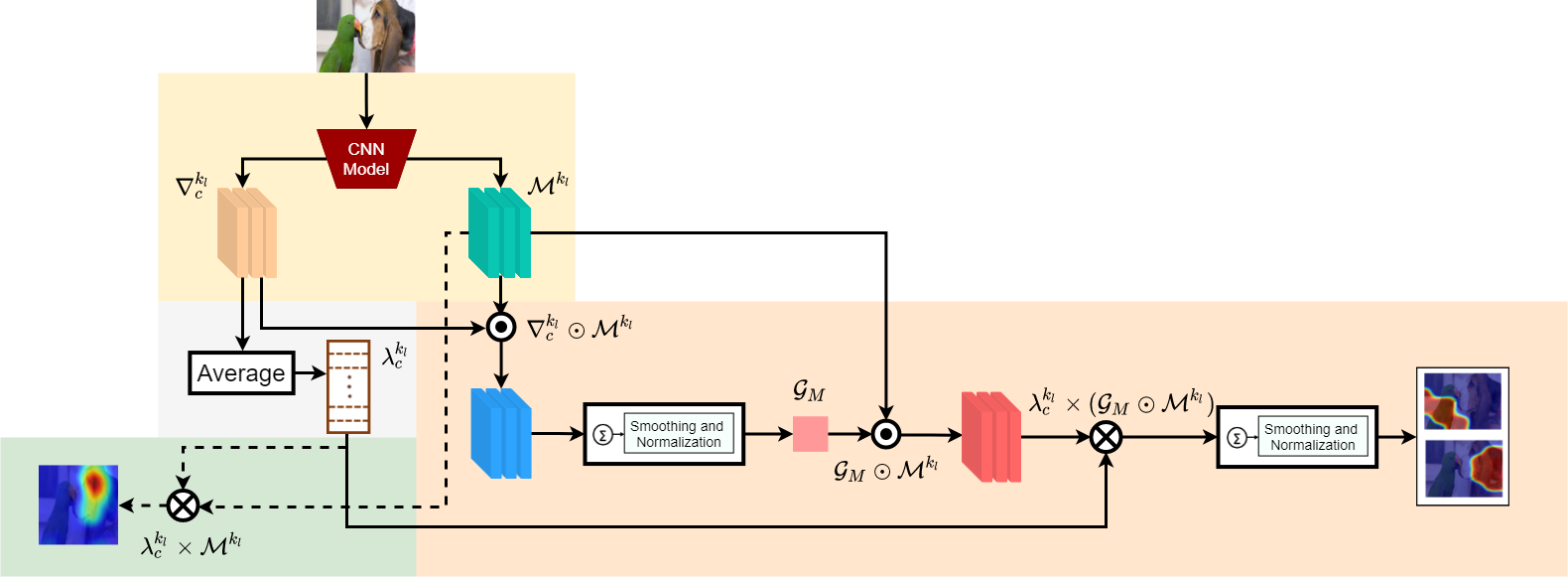}
	\caption{Visual comparison between previous state-of-the-art studies and the proposed method. For the given demonstrations, our approach can mark down the primary salient regions under challenging visual conditions. Additionally, our saliency maps are more concrete and leaves almost no traces for the secondary-salient areas.}
	\label{architecture}
\end{figure*}

This section will describe the formulation of the proposed approach as shown in figure \ref{architecture} for obtaining the saliency for the given model and image.

\subsection{Baseline formulation }
Every trained model infers through the collective response of its feature maps for the given image. The current understanding of deep learning requires more research to define the ideal formulation for the extracted feature maps during inference. During inference, activated feature maps can be sparse, shallow, and rich in containing the spatial correspondence and their collective distribution governs the outcome of the final prediction. 

Let $\mathcal{\phi}$ be the model we take for the inference. For any image $\mathcal{X} \in \mathcal{R}^{W \times H \times 3}$, model $\mathcal{\phi}$ generates feature maps $\mathcal{M}$. Say, $\mathcal{M}^{k_l}$ is the $k^{th}$ feature channel at $l^{th}$ layer, before the final dense layer. If we aggregate them all, we will obtain a unified map that will represent the collective correspondence due to the model $\mathcal{\phi}$ for the given image. This aggregation for the activated channels is as follows: 

\begin{equation}\label{eq1}
   \mathcal{A}^{l} = \sum_{k} \mathcal{M}^{k_l}
\end{equation}

This tensor $\mathcal{A}^{l}$ contains the global representation for all activated feature maps, which can serve the purpose of marking salient regions with careful tuning as shown in figures \ref{sinef}, and \ref{duof}. In equation (\ref{eq1}),  we aggregated all of the feature representations into $\mathcal{A}^{l}$; elements over a specific threshold may correspond to the primary class information. Nonetheless, this observation only works within images with a single class, but is not appropriate for the dual class scenarios, as shown in figure \ref{algorithm figure}.

To achieve class discriminatory behaviour, researchers \cite{Grad-CAM,Grad-CAM++,integrated,FullGrad} worked with the idea of using weighted aggregation for equation (\ref{eq1}), in contrast to plain linear addition operation. To extract weights, first they compute the gradient maps  $\nabla_\mathcal{C}$ with respect to the given class $\mathcal{C}$, from the feature maps $\mathcal{M}$. 

If $Y^{\mathcal{C}}$ is the class score \cite{Grad-CAM} for the given image from the input model $\mathcal{\phi}$, for each location $(i,j)$ of the $k^{th}$ feature map at the $l^{th}$ layer, then the corresponding gradient map is expressed as :

\begin{equation}\label{eq2}
   \nabla_{\mathcal{C}_{i j}}^{k_{l}} =  \frac{\partial Y^{\mathcal{C}}}{\partial \mathcal{M}_{i j}^{k_{l}}} 
\end{equation}

If each feature map holds $\mathcal{Z}$ number of elements, then the corresponding weight for each feature map $\mathcal{M}^{k_l}$ is:

\begin{equation}\label{eq3}
     \lambda_{\mathcal{C}}^{k_{l}}= \frac{1}{\mathcal{Z}}\sum_{i} \sum_{j} \nabla_{\mathcal{C}_{i j}}^{k_{l}} 
\end{equation}

which is the mean value of $\nabla_{\mathcal{C}}^{k_{l}}$. Hence, the regular baseline formulation \cite{Grad-CAM} for the saliency map  $\mathcal{S}_{\mathcal{C}}$ estimation is expressed as follows:

\begin{equation}\label{eq4}
    \mathcal{S}_{\mathcal{C}} =  \textit{ReLU}({\sum_{k}    \lambda_{\mathcal{C}}^{k_{l}}\times \mathcal{M}^{k_l}})
\end{equation}

\subsection{Incorporating global guidance } 
Equation (\ref{eq4}) achieves class discriminatory behavior, however it still faces challenges due to its formulation. If we investigate the above equation, we see that $\lambda_{\mathcal{C}}^{k_{l}}$  weighs the corresponding feature map $\mathcal{M}^{k_l}$. Typical $\lambda_{\mathcal{C}}^{k_{l}}$ treats every member of the given  $\mathcal{M}^{k_l}$ equally and increases/decreases their collective effect homogeneously. Therefore, we can still see the traces of the other classes in the saliency map, as shown in figure \ref{algorithm figure}. Other studies \cite{SISE_main,Ada-SISE} have shown class discriminatory performance through perturbations without relying upon the gradients. However, those studies are time expensive \cite{FullGrad} and often require human interaction. Hence, it is natural to ask, can we still rely on the gradient-weighted operation and address the above issues? 

In response to this, we propose global guidance map. To obtain the global guidance map, we perform a simple elementwise multiplication between the feature maps and their corresponding gradient maps. The formulation for the global guidance map is as follows:

\begin{equation}\label{eq5}
    \mathcal{G}_{M} =  \textit{ReLU}({\sum_{k}    \nabla_{\mathcal{C}}^{k_{l}}\odot \mathcal{M}^{k_l}})
\end{equation}

\begin{figure*}[!htbp]
\centering
\begin{subfigure}[l]{0.18\textwidth}
    \centering
    \includegraphics[height=24mm,width=30mm]{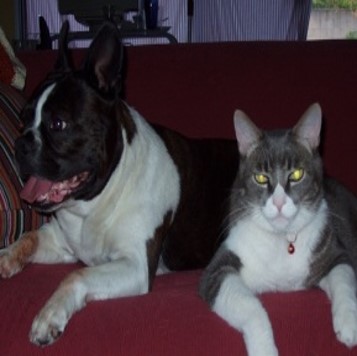}
    \subcaption{Input}
    \label{fig:a}
    \centering
\end{subfigure}
\begin{subfigure}[r]{0.70\textwidth}
    \centering
    \begin{subfigure}[t]{0.25\textwidth}
        \centering
        \includegraphics[height=22mm,width=\textwidth]{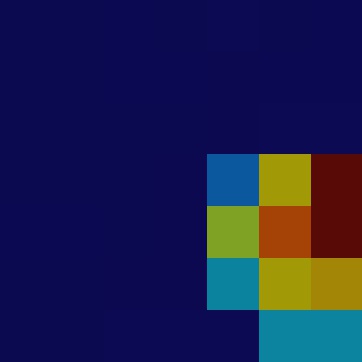}
       \subcaption{$g_M$(cat)}
        \label{fig:b}
    \end{subfigure}
        \begin{subfigure}[t]{0.25\textwidth}
        \centering
        \includegraphics[height=22mm,width=\textwidth]{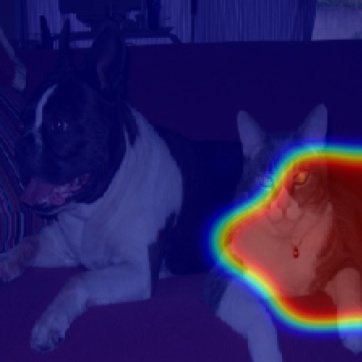}\\
        \subcaption{$S_c$ w/o $g_M$(dog)}
        \label{fig:c}
    \end{subfigure}
     \begin{subfigure}[t]{0.25\textwidth}
        \centering
        \includegraphics[height=22mm,width=\textwidth]{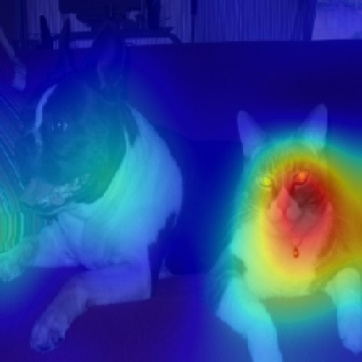}\\
        \caption{$S_C$ w/o $g_M$ for \cite{Grad-CAM++}(cat)}
        \label{fig:d}
    \end{subfigure}\\
    \begin{subfigure}[t]{0.25\textwidth}
        \centering
        \includegraphics[height=22mm,width=\textwidth]{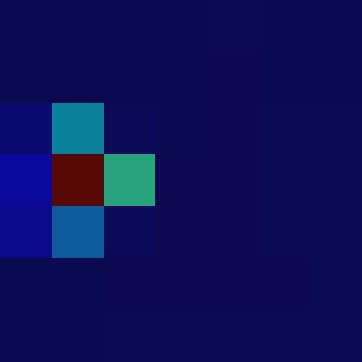}\\
        \caption{$g_M$ (dog)}
        \label{fig:e}
    \end{subfigure}
        \begin{subfigure}[t]{0.25\textwidth}
        \centering
        \includegraphics[height=22mm,width=\textwidth]{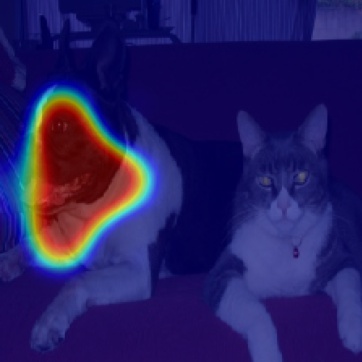}\\
        \caption{$S_C$ w/o $g_M$(dog)}
        \label{fig:f}
    \end{subfigure}
        \begin{subfigure}[t]{0.25\textwidth}
        \centering
        \includegraphics[height=22mm,width=\textwidth]{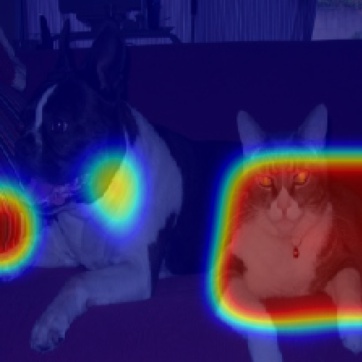}\\
        \caption{$S_C$ w/o $g_M$(dog)}
        \label{fig:g}
    \end{subfigure}
        \begin{subfigure}[t]{0.25\textwidth}
        \centering
        \includegraphics[height=22mm,width=\textwidth]{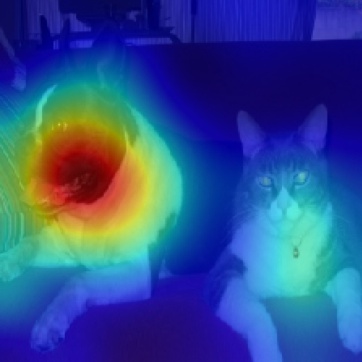}\\
        \caption{$S_C$ w/o $g_M$ for \cite{Grad-CAM++}(dog)}
        \label{fig:h}
    \end{subfigure}
     
\end{subfigure}
\caption{Proposed global guidance maps $g_M$ for cat, proposed saliency map ($S_c$) with and without global guidance
(first row), and the same maps for the dog [second row and third row]. The proposed global guidance map provides strong localization
information as well as exclusion of non-target class.}
\label{rebuttal}
\end{figure*}

The idea behind considering the global guidance map is to focus only on the salient regions by limiting the operating zone for the $\lambda_{\mathcal{C}}^{k_{l}}$. Since the multiplication operation  of equation (\ref{eq5}) investigates individual responses for every element of the gradient maps, we can mark the class-specific regions from the feature maps. In this way, captured guidance map successfully omits the trace of the secondary classes from equation (\ref{eq4}) unless the generated feature maps heavily overlap between categories, signifying the possible misclassification. 

In summary, we first compute $\mathcal{G}_{M}$ and multiply it with each of the feature maps to obtain the class discriminative feature maps from the initial class  representative feature maps. Then, we perform the weighted multiplication between each member of the $\lambda_{\mathcal{C}}^{k_{l}}$ and class discriminative feature maps, followed by the final aggregation to obtain the desired representation. Hence, our proposed weighted-multiplicative-aggregation is as follows:

\begin{equation}\label{eq6}
   \mathcal{S}_{\mathcal{C}} =  \textit{ReLU}({\sum_{k}    \lambda_{\mathcal{C}}^{k_{l}}\times (\mathcal{G}_{M}\odot\mathcal{M}^{k_l}}))
\end{equation}
 
In equation (\ref{eq6}), the only reason we are using the $\lambda_{\mathcal{C}}^{k_{l}}$ is to gain a similar homogeneous increment as in equation (\ref{eq4}), which also helps to preserve visual integrity during the final upsampling operation. Since guidance map $\mathcal{G}_{M}$ can successfully omit the secondary classes from any $k^{th}$ feature map $\mathcal{M}^{k_l}$ by \lq masking\rq\space the primary governing region, it also becomes possible to produce increments in the desired class region with the additional multiplication help from $\lambda_{\mathcal{C}}^{k_{l}}$. Finally, typical smoothing and normalization are performed on the saliency map before post aggregation up-sampling onto the given image size. In this way, the achieved saliency map $\mathcal{S}_{\mathcal{C}}$ shows significant improvement in both single class representative and multiclass discriminative cases.

\begin{figure*}[!htbp]
	\centering
	\includegraphics[width = 5.8 in, height= 3.3 in]{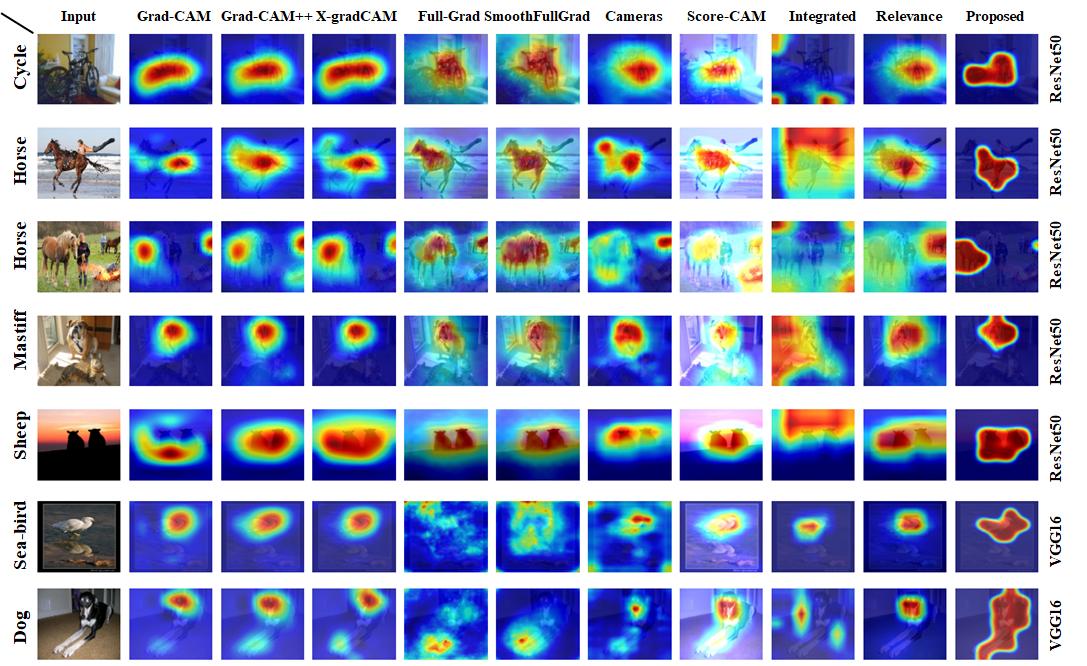}
	\caption{Visual comparison between previous state-of-the-art studies and the proposed method. For the given demonstrations, our approach can mark down the primary salient regions under challenging visual conditions. Additionally, our saliency maps are more concrete and leaves almost no traces for the secondary-salient areas.}
	\label{qualitative result}
\end{figure*}

\section{Performance evaluation}
\textbf{Datasets.} 
Our experimental setup covers three widely used vision datasets: ImageNet, MS-COCO 14, and PASCAL-VOC 12. Among them, PASCAL-VOC 12 provides the full segmentation annotations for the input images. Hence, experiments using segmentation-oriented metrics are from this dataset. Other experiments do not involve segmentation labels, as the rest of the experiments are applicable for all datasets mentioned above. For the ImageNet and MS-COCO 14 datasets, we randomly selected a few thousand \cite{Score-CAM,Grad-CAM++,Axiom-CAM} for the experiments.

\textbf{Compared studies.} 
For visual comparison, given the availability and functional complexities, we consider available official implementations of the GradCAM\cite{Grad-CAM}, GradCAM++\cite{Grad-CAM++}, X-gradCAM\cite{Axiom-CAM}, ScoreCAM\cite{Score-CAM}, Fullgrad\cite{FullGrad}, Smooth-Fullgrad\cite{FullGrad}, CAMERAS\cite{Cameras}, Integrated\cite{integrated}, and Relevance-CAM\cite{relevance}. We report seven different quantitative analyses in addition to the visual demonstration. We  extract saliency maps from pretrained VGG16 \cite{VGG16} and ResNet50\cite{ResNet50} networks for all of the methods above in our experiments. A later section reports the quantitative results from ResNet50 in the respective tables. As with previous studies \cite{Score-CAM,integrated,Grad-CAM++,Axiom-CAM}, few thousand random images are selected from the datasets for our experiments. However, since the random images from previous studies are not made public, the results of our random selections may vary from the cited studies. %Our selections of images are presented in the supplementary materials along with the code for the proposed method. 

\begin{figure*}[htbp]
\begin{center}
  \begin{subfigure}{0.7\textwidth}
    \includegraphics[width=5in,height=1.8in]{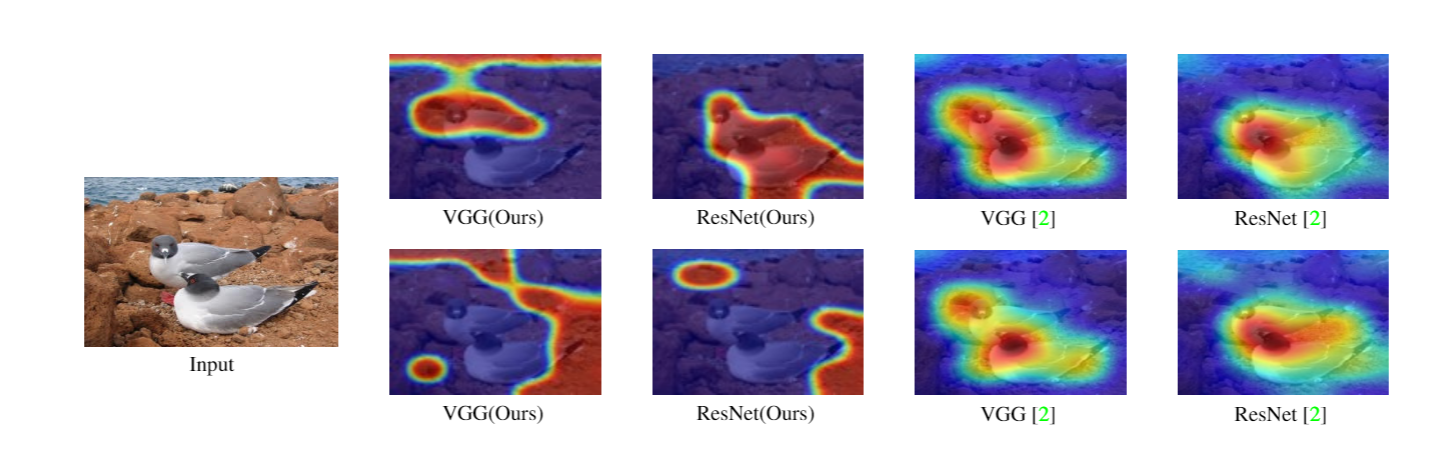}
    \caption{ Interpretation comparison between proposed and GradCAM++\cite{Grad-CAM++}}
    \label{4.1}
  \end{subfigure}
\end{center}
\begin{center}
  \begin{subfigure}{0.7\textwidth}
    \includegraphics[width=5in,height=0.6in]{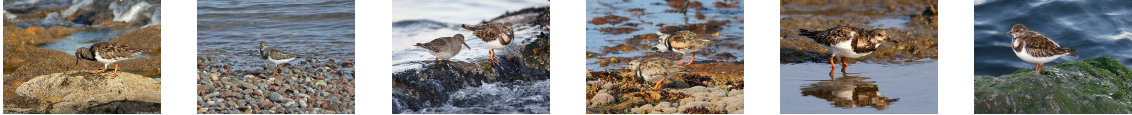}\\
   \caption{``Ruddy Turnstone" images from the ``ImageNet" dataset.}
    \label{4.2}
  \end{subfigure}
\end{center}
  \caption{Here, (a) shows the interpretation comparison between the proposed and GradCAM++\cite{Grad-CAM++}. The upper row is the response map for the primary ``Albatross" bird class, and the second row is for the secondary class, ``Ruddy Turnstone" bird. The proposed method clearly presents the difference between VGG16 and ResNet50 in terms of interpretation, whereas GradCAM++ responses are all similar across different networks and classes. 
  (b) As we suspect that dataset bias might lead to such a decision for the given ``Albatross" images, an inspection of the ImageNet dataset could clarify further. Upon examining typical Ruddy Turnstone bird images in the ImageNet dataset, we see that stony shore is the background for most of the Ruddy Turnstone bird images. }
  \label{4}
\end{figure*}

\subsection{Visual demonstration}
In this section, we show a qualitative comparison with previous methods. In figure \ref{qualitative result}, we have included diverse sets of images from the datasets. The first five rows show the performance due to ResNet50, and later are for VGG16. Our image selection includes class representative, class discriminative, and multiple instances examples. Here, the proposed method can capture the class discriminative region with greater confidence, if not the entire class area itself for the bicycle, sheep, sea-bird, and dog image. Our scheme captures the class region for the sea-bird image and excludes its reflection in the water compared to other methods.

Additionally, the proposed CAM bounds the whole dog class as a dog and captures the sheep in a lowlight environment more clearly than other methods. For images with dual classes, our method presents superior class discriminative performance. Our study successfully bounds the horse regions in the horse images without leaving any traces to other class regions. On the other hand, compared studies often mark both horse and secondary class instances.

\subsection{Quantitative analysis}
To present our quantitative analysis, we perform the following experiments: model's performance drops and increments due the to salient and context regions, Pointing score, Dice score, and IoU score. We present the scores for the above metrics for ResNet50 for all datasets.

\textbf{Performance due to the saliency region.} 
If we have a perfect model and a perfect interpreter to mark the spatial correspondence for the specific class, the network will provide a similar prediction for both the given image and the segmented salient image. Here, we first extract the salient region from the given images with the help of the given interpreter. Then we perform prediction on the original image, and the corresponding salient image \cite{Grad-CAM++} and check the performance drop for the given interpreter. The expectation is that the better interpreter can exclude the non-salient region as much as possible; hence, the performance drop will be as low as possible. Therefore, our first metric delivers the performance drop due to salient area only as input. For some cases, prediction performance hinders due to the presence of strong spatial context. 
%So, if the network can predict the correct class within such context with $p1$ confidence, and prediction without such context is $p2$, then $p1 < p2$. This metric will give us the score for performance increment due to the salient zone only.

\begin{table*}[!htbp]
\centering
\caption{Comparative evaluation in terms of salience zone drop and context zone increase, Pointing Game, Dice, IoU (higher is better) and salience zone increase and context zone drop (lower is better) on the PASCAL VOC 2012 dataset for ResNet50 model.
The best scores are in bold form and second best scores are  underlined.}
\resizebox{\textwidth}{!}{
\begin{tabular}{lcccccccccc}
\hline
                                 & GradCAM\cite{Grad-CAM} & GradCAM++\cite{Grad-CAM++} & X GradCAM\cite{Axiom-CAM} & CAMERAS\cite{Cameras} & FullGrad\cite{FullGrad} & SmoothFullGrad\cite{FullGrad} & Integrated\cite{integrated} & ScoreCAM\cite{Score-CAM} 
                            & Relevance\cite{relevance}
                            & Proposed \\
\hline

Increase for salience zone $\uparrow $     & 0.0366 & \underline{0.0716} & 0.0366 & 0.0715 & 0.0421 & 0.0421 & 0.0172 & 0.0506 & 0.0559&\textbf{0.0786} \\
Drop for context zone $\uparrow$           & 0.9395 & 0.8812 & \underline{0.9429} & 0.8834 & 0.9083 & 0.9281 & 0.8124 & 0.9389 & 0.9089&\textbf{0.9443} \\
Pointing Game $\uparrow$                   & 0.3355 & 0.4731 & 0.3733 & 0.4412 & 0.3713 & 0.4422 & 0.2642 & 0.4322& \underline{0.532}&\textbf{0.5945}  \\
Dice  $\uparrow$                           & 0.2822 & 0.3422 & 0.2934 & 0.3342 & 0.2942 & 0.3328 & 0.1834 & 0.3321 & \underline{0.411}&\textbf{0.4321} \\
IoU   $\uparrow$                         & 0.0823 & 0.1122 & 0.0901 & 0.1007 & 0.0812 & 0.0963 & 0.0625 & 0.0943 & \underline{0.121}&\textbf{0.1321} \\
Drop for salience zone $\downarrow$        & 0.8996 & \underline{0.8064} & 0.8745 & 0.8201 & 0.8873 & 0.8762 & 0.9399 & 0.8567 & 0.8333&\textbf{0.7784} \\
Increase for context zone $\downarrow $    & 0.0172 & 0.0312 & 0.0205 & 0.0244 & 0.0291 & 0.0215 & 0.0411 & \textbf{0.0151} &0.029& \underline{0.0183} \\

\hline
\end{tabular}
}
\centering

\label{t1}
\end{table*}

\begin{table*}[!htbp]

\caption{ Comparative performance drop and increment of saliency and context zones on the MS-COCO 14 dataset for ResNet50 model.}

\centering

\resizebox{\textwidth}{!}{
\begin{tabular}{lcccccccccc}
\hline
                                        
                                          & GradCAM\cite{Grad-CAM} & GradCAM++\cite{Grad-CAM++} & X GradCAM\cite{Axiom-CAM} & CAMERAS\cite{Cameras} & FullGrad\cite{FullGrad} & SmoothFullGrad\cite{FullGrad} & Integrated\cite{integrated} & ScoreCAM\cite{Score-CAM} &
                                          Relevance\cite{relevance}&                           Proposed \\
\hline
Drop for context zone $\uparrow$            & 0.9023 & \textbf{0.9495} & 0.9142 & \underline{0.9424} & 0.8961 & 0.9025 & 0.8607 & 0.9183 & 0.9081 & 0.9391 \\
Increase for saliency zone $\uparrow$       & 0.0490 & \underline{0.0913} & 0.0555 & 0.0935 & 0.0455 & 0.0495 & 0.0796 & 0.0695 & 0.089  & \textbf{0.0999} \\
Drop for saliency zone $\downarrow$         & 0.8394 & \underline{0.7243} & 0.8081 & 0.7325 & 0.8454 & 0.8288 & 0.7713 & 0.7822 & 0.7357 & \textbf{0.6493} \\
Increase for context zone $\downarrow$      & 0.0245 & \underline{0.0165} & 0.0215 & 0.0172 & 0.0311 & 0.0315 & 0.0335 & 0.0205 & 0.0311 & \textbf{0.0152} \\
\hline
\end{tabular}
}
\centering

\label{t2}

\end{table*}

\begin{table*}[!htbp]
\caption{ Comparative performance drop and increment of saliency and context zones on the ImageNet
dataset for ResNet50 model.}
\centering

\resizebox{\textwidth}{!}{
\begin{tabular}{c c c c c c c c c c c}
\hline
                                        
                                          & GradCAM\cite{Grad-CAM} & GradCAM++\cite{Grad-CAM++} & X GradCAM\cite{Axiom-CAM} & CAMERAS\cite{Cameras} & FullGrad\cite{FullGrad} & SmoothFullGrad\cite{FullGrad} & Integrated\cite{integrated} & ScoreCAM\cite{Score-CAM} & 
                                    Relevance\cite{relevance}&      Proposed \\
\hline
Drop for context zone $\uparrow$            & 0.8767 & 0.9178 & 0.8698 &\underline{0.9335} & 0.8379 & 0.8548 & 0.8938 & 0.8585 & 0.8541 &\textbf{0.9392} \\
Increase for saliency zone $\uparrow$       & 0.0535 & 0.0903 & 0.0635 & 0.0936 & 0.0803 & 0.0669 & 0.0435 & \underline{0.1003} & 0.0969 &\textbf{0.1008} \\
Drop for saliency zone $\downarrow$        & 0.7906 & \underline{0.6717} & 0.7682 & 0.7302 & 0.7775 & 0.7844 & 0.7267 & 0.6975 & 0.6844 &\textbf{0.6492} \\
Increase for context zone $\downarrow$      & 0.0234 & \underline{0.0067} & 0.0368 & 0.0134 & 0.0502 & 0.0301 & 0.0301 & 0.0468 & 0.0602 &\textbf{0.0012} \\
\hline
\end{tabular}
}
\centering

\label{t3}

\end{table*}

\textbf{Performance due to the context region.} 
As above, if saliency extraction is as good as one expects, then we can set up another experiment with the context region. In this setup, we first exclude the salient part from the given image to obtain the context image and predict it. If the interpreter can successfully extract all the salient areas, the performance will drop near 100 percent. %Conversely, networks can predict from contexts, which in few cases, prediction can increase; however, this is very unlikely. 

\textbf{Pointing game, Dice Score, IoU.} 
For image sets with segmenation labels, various segmentation evaluations can be calculated for saliency maps. We follow \cite{pointinggame} to perform the pointing game for class discriminative evaluation. In this performance metric, the ground-truth label is used to trigger each visualization approach, and the maximum active spot on the resulting heatmap is extracted. After that, it determines if the highest saliency points undergo into the annotated boundary box of an object, determining whether it is a hit or miss. The term \begin{tiny}$\frac{ Hits_{total}}{Hits_{total}+ Misses_{total}}$\end{tiny} is calculated as the pointing game accuracy, which high value represents a better explanation for any model. Dice score is a popular metric for analyzing segmentation performance. It results from the ratio of the doubled intersection over the total number of elements associated with this instance. IoU stands for intersection over the union. It is a widespread metric for evaluating segmentation performance. This score ranges from 0 to 1 and signifies the overlapping area between the obtained image and its corresponding ground truth.
 
%%\vspace{-1.3cm}
\begin{figure*} [htbp]
\centering
\begin{tabular}{cccccc}

\includegraphics[width=0.8in,height =0.8 in]{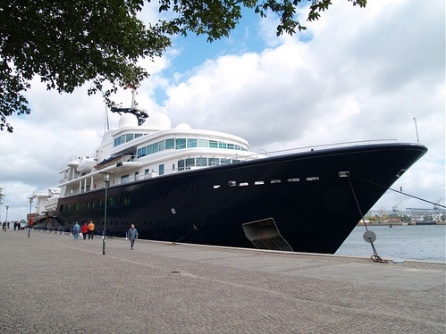}& 
\includegraphics[width=1.8in,height =1.2 in]{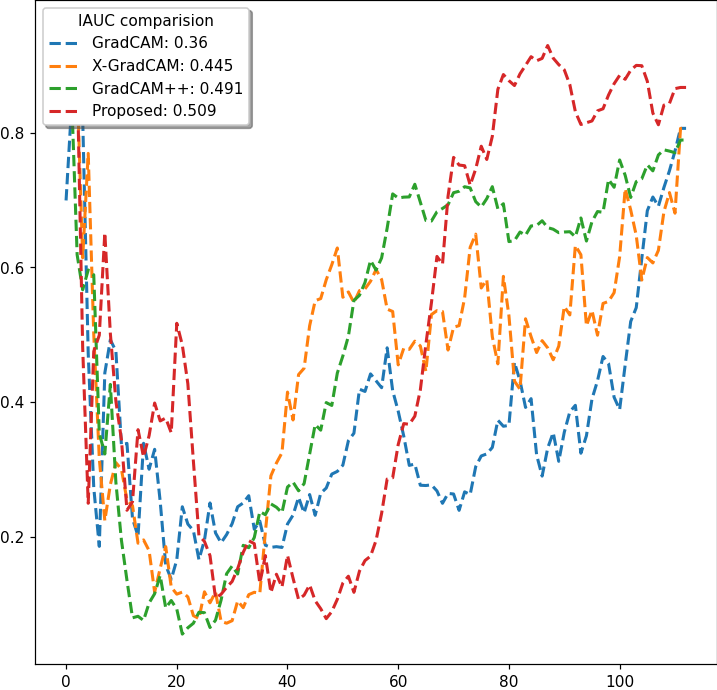} &  
\includegraphics[width=1.8in,height =1.2 in]{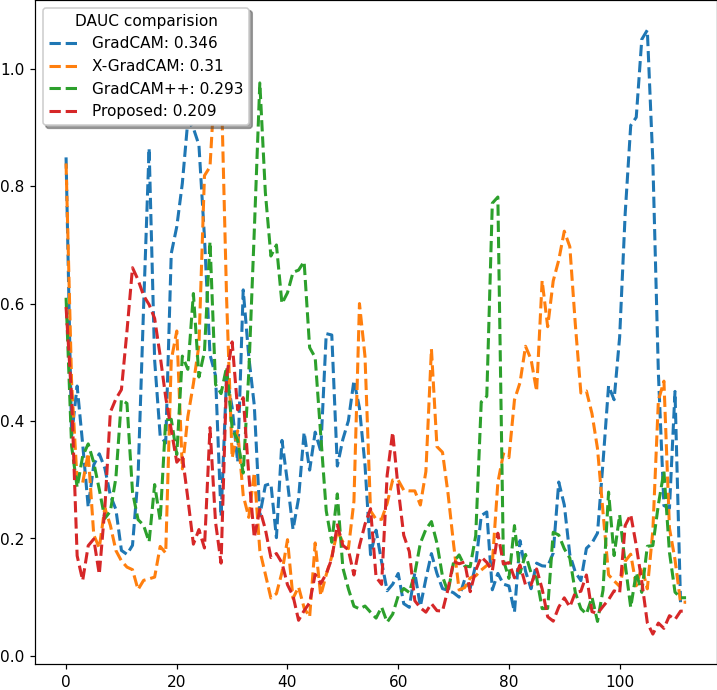} \\
\tiny 5.1(a)  & \tiny 5.1(b) & \tiny 5.1(c)  \\[1pt]
\end{tabular}
\\
\begin{tabular}{cccc}
\includegraphics[width=0.8in,height =0.8in]{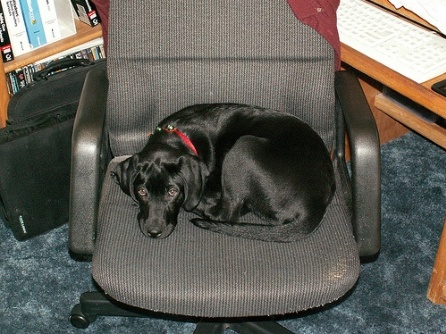} &
\includegraphics[width=1.8 in,height =1.2 in]{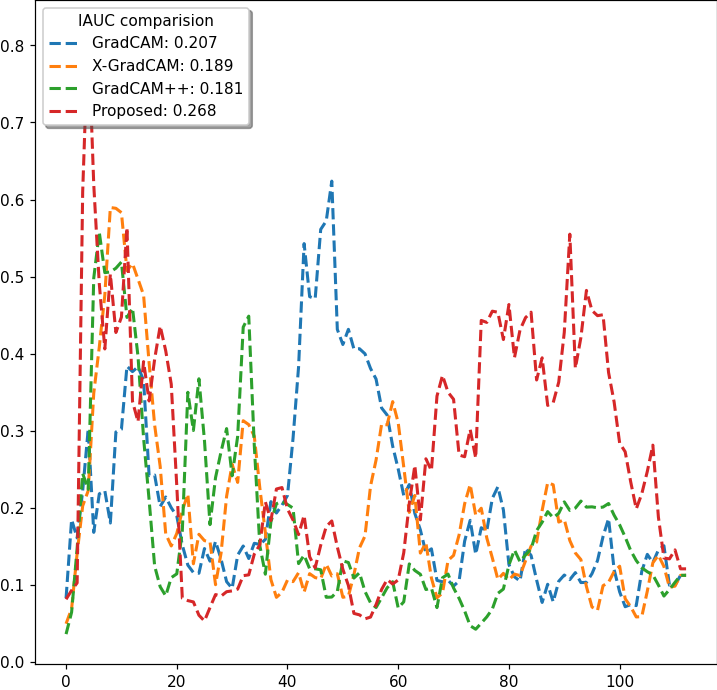} &
\includegraphics[width=1.8 in,height =1.2 in]{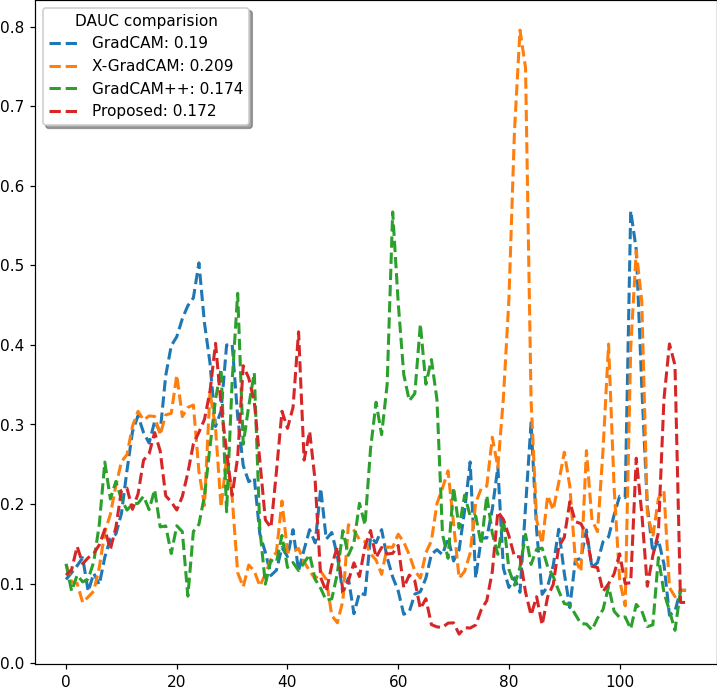} \\
\tiny 5.2(d)  & \tiny 5.2(e)  & \tiny 5.2(f)\\[1pt]
\end{tabular}
\\
\begin{tabular}{cccc}
\includegraphics[width=0.8in,height =0.8in]{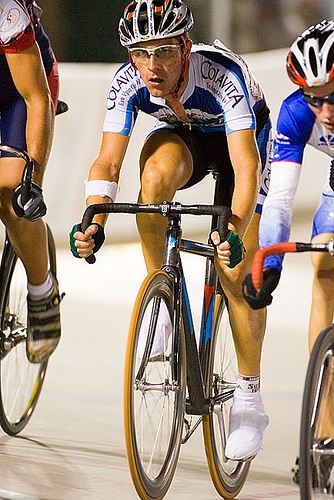} &
\includegraphics[width=1.8 in,height =1.2 in]{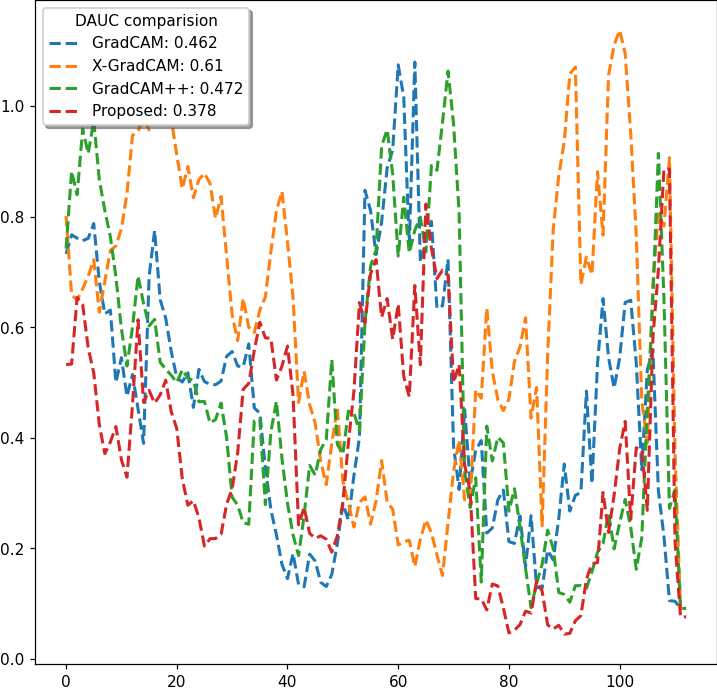} &
\includegraphics[width=1.8 in,height =1.2 in]{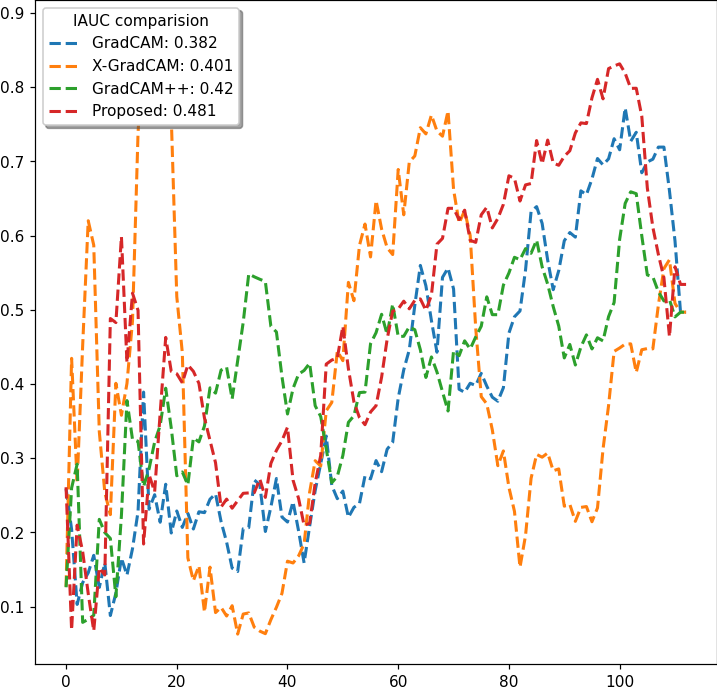} \\
\tiny 5.3(g)  & \tiny 5.3(h)  & \tiny 5.3(i)\\[1pt] 

\end{tabular}

\caption{ AUC demonstration for the insertion and deletion operation for images on the left side. The above analysis shows that the proposed method can capture the most salient regions for a single class, dual-class, and dual-class with multiple instances compared to the previous methods \cite{Grad-CAM,Axiom-CAM,Grad-CAM++}.}

\label{fig:AUC2} 
\end{figure*}

The proposed method performs better or similar to the best performing saliency generation method in table \ref{t1}.  Here, we present the comparative data on the Pascal VOC 2012 dataset for the ResNet50 model. Out of seven different performance tests, our method obtains the highest score for the six of them. For an increase in the context zone, our study differs only 0.03 from the best performing result. In table \ref{t2}, the proposed research avails state-of-the-art performance for three out of four metrics. Table \ref{t3} also shows the best performance of our method for every experiment metrics using the ImageNet dataset. The PASCAL VOC 2012 dataset has corresponding segmentation mask, and we can achieve pseudo segmentation masks by thresholding the saliency maps. Achieved scores for the Pointing game, Dice and IoU signify that our study can better capture interest zones than the compared studies. To measure the explainability, we also have conducted a comparative analysis in figure \ref{fig:AUC2} on three images from the Pascal VOC 2012 dataset and presented the insertion and deletion operation.  Here, our method captures the most salient regions for both single, dual, and dual classes with multiple instances in comparison to GradCAM \cite{Grad-CAM}, Grad-CAM++\cite{Grad-CAM++}, and X-GradCAM\cite{Axiom-CAM}. %We have included more results from our research in the supplementary material. 

\subsection{Interpretation comparison}
Saliency map generation is not all about capturing the class of interest as precisely as possible. A faithful interpretation is also a significant part of saliency generation studies. In other words, any interpretable method should explain why the underlying model is making such a prediction by marking the corresponding image region. However, we cannot present this for every image from the dataset, but a sophisticated example can show the difference from previous methods.

In figure \ref{4.1}, we have presented the interpretation comparison between the proposed study and the GradCAM++\cite{Grad-CAM++}. Here, the top-1 class response is \lq Albatross\rq\space for the given image. For VGG16, saliency map for the proposed method marks one of the Albatross birds and the water as context, but fails to mark other Albatross bird. For \lq Albatross\rq, our method with the VGG16 shows different interpretations due to the global guidance map and responds to the water as context. In contrast to ResNet50, our guidance map marks both Albatross birds without marking the water context. However, for both networks, GradCAM++\cite{Grad-CAM++} captures both of the birds also barely touches the water context. For this particular image, the proposed method presents clearer interpretation difference between VGG16 and ResNet50. 

In \ref{4.2}, we show why the models might identify the given image as \lq Ruddy Turnstone\rq\space class. With our scheme, we interpret that surrounding stones and water are features that are corresponding to Ruddy Turnstone bird for both VGG16 and ResNet50. This interpretation makes more sense if we look at typical Ruddy Turnstone images from the ImageNet dataset \ref{4.2}, where the most of Ruddy Turstone birds are shown with stony sea-shore area as the background. Hence, we can utilize this interpretation as a medium for identifying the dataset bias. On the other hand, GradCAM++\cite{Grad-CAM++} shows the Albatross bird regions as the interpretation for the Ruddy Turnstone, and almost ignores the associated context.

%Practically  ; this comparison is self-explanatory for why the proposed interpretation scheme is more reliable.
%In support of our claim, we refer to figure \ref{4.2}, where more images are present for the Albatross images. This figure presents the saliency map for the "Albatross" and the "Ruddy Turnstone" classes from Albatross images, in addition to the accuracy transition for pixel insertion and deletion. As we state, water and stone context is the critical factor in deciding specific predictions for the Albatross images; it is present in the insertion and deletion curve analysis. Figure \ref{4.2} further bolsters our claim, as randomly chosen photos for the "Ruddy Turnstone" class from the ImageNet dataset. From this figure, all of the images contain the "Ruddy Turnstone"  subject with a similar background where water and stony regions are present.

%However, our method also faces some challenges within its explanatory purpose. If the given image contains multiple instances of the same class, the proposed method is less likely to emphasize each instance. This particular trend majorly comes from the inherent network's feature maps. A typical example of such a scenario is present in the supplementary material. We would like to investigate layer-wise relevance propagation or rectified gradient inclusion concepts to address our current limitations in our future work.

\section{Conclusion}
In this study, we present a novel extension of the traditional gradient-dependent saliency map generation scheme. The proposed method leverages element-wise multiplicative aggregation as guidance with previous weighted multiplicative summation and further improves the performance of salient region bounding. Additionally, we showed our study's advanced class discriminative performance and presented evidence for better area framing with deeper networks. Furthermore, our model can avail crisper saliency map and significant quantitative improvement over three widely used datasets. We aim to integrate this study into other vision tasks in future work.

%Bibliography
\bibliographystyle{unsrt}  
\bibliography{references}

\end{document}